\documentclass[sigconf,nonacm]{acmart}
\usepackage{amssymb}
\usepackage{graphicx}  % 处理图片（仅加载1次）
\usepackage{amsmath, amssymb}  % 数学公式与符号（合并加载，仅1次）
\usepackage{bm}  % 粗体数学符号（如 \bm{\beta}）

\usepackage{booktabs}  % 专业表格线（仅加载1次）
\usepackage{multirow}  % 跨多行单元格
\usepackage{array}  % 增强表格列格式控制
\usepackage{makecell}  % 单元格内换行/对齐
\usepackage{colortbl}  % 表格颜色（按需保留，不用可删除）

\usepackage{appendix}
\AtBeginDocument{%
  }

\begin{document}

%%
%% The "title" command has an optional parameter,
%% allowing the author to define a "short title" to be used in page headers.
\title{HGMF: A Hierarchical Gaussian Mixture Framework for Scalable Tool Invocation within the Model Context Protocol}

%%
%% The "author" command and its associated commands are used to define
%% the authors and their affiliations.
%% Of note is the shared affiliation of the first two authors, and the
%% "authornote" and "authornotemark" commands
%% used to denote shared contribution to the research.
\author{Wenpeng Xing}
\authornote{Both authors contributed equally to this research.}
\affiliation{%
  \institution{Zhejiang University \\ Binjiang Institute of Zhejiang University}
  \city{Hangzhou}
  \country{China}}
\email{wpxing@zju.edu.cn}
\orcid{1234-5678-9012}

\author{Zhipeng Chen}
\authornotemark[1]
\affiliation{%
  \institution{Jimei University}
  \city{Xiamen}
  \country{China}
}
\email{zhipengchen@jmu.edu.cn}
\orcid{1234-5678-9012}

\author{Changting Lin}
\email{linchangting@gmail.com}
\author{Meng Han}
\authornote{Corresponding Author.}
\email{mhan@zju.edu.cn}
\affiliation{%
  \institution{Zhejiang University \quad GenTel.io}
  \city{Hangzhou}
  \country{China}}

% \author{Meng Han}
% \affiliation{%
%   \institution{Zhejiang University \quad GenTel.io}
%   \city{Hangzhou}
%   \country{China}
% }\email{mhan@zju.edu.cn}

%%
%% By default, the full list of authors will be used in the page
%% headers. Often, this list is too long, and will overlap
%% other information printed in the page headers. This command allows
%% the author to define a more concise list
%% of authors' names for this purpose.
\renewcommand{\shortauthors}{Trovato et al.}

%%
%% The abstract is a short summary of the work to be presented in the
%% article.
\begin{abstract}
Invoking external tools enables Large Language Models (LLMs) to perform complex, real-world tasks, yet selecting the correct tool from large, hierarchically-structured libraries remains a significant challenge. The limited context windows of LLMs and noise from irrelevant options often lead to low selection accuracy and high computational costs. To address this, we propose the Hierarchical Gaussian Mixture Framework (HGMF), a probabilistic pruning method for scalable tool invocation. HGMF first maps the user query and all tool descriptions into a unified semantic space. The framework then operates in two stages: it clusters servers using a Gaussian Mixture Model (GMM) and filters them based on the query's likelihood. Subsequently, it applies the same GMM-based clustering and filtering to the tools associated with the selected servers. This hierarchical process produces a compact, high-relevance candidate set, simplifying the final selection task for the LLM. Experiments on a public dataset show that HGMF significantly improves tool selection accuracy while reducing inference latency, confirming the framework's scalability and effectiveness for large-scale tool libraries.
\end{abstract}

%%
%% The code below is generated by the tool at http://dl.acm.org/ccs.cfm.
%% Please copy and paste the code instead of the example below.
%%


%%
%% Keywords. The author(s) should pick words that accurately describe
%% the work being presented. Separate the keywords with commas.
%% A "teaser" image appears between the author and affiliation
%% information and the body of the document, and typically spans the
%% page.

%\received{20 February 2007}
%\received[revised]{12 March 2009}
%\received[accepted]{5 June 2009}

%%
%% This command processes the author and affiliation and title
%% information and builds the first part of the formatted document.

\begin{abstract}

Invoking external tools empowers large language models (LLMs) to execute complex, real-world tasks, yet selecting the appropriate tool from expansive, hierarchically structured libraries poses a formidable challenge. LLM context limitations and semantic noise from irrelevant options frequently yield suboptimal accuracy and elevated computational overhead. To mitigate these issues, we introduce the Hierarchical Gaussian Mixture Framework (HGMF), a probabilistic pruning paradigm for scalable tool invocation under the Model Context Protocol. HGMF begins by embedding user queries, servers, and tools into a unified semantic space via sentence transformers with L2 normalization. It then employs a two-stage GMM-based filtering: server-level clustering identifies query-relevant clusters via likelihood maximization, pruning to a subset; tool-level GMMs, applied within selected servers, refine candidates. To enhance robustness in low-sample regimes---where cluster boundaries blur and distributions destabilize---we integrate inter-class regularization (promoting separation between cluster centers) and intra-class regularization (ensuring compactness within clusters), balanced with covariance stabilization. The resultant compact, high-fidelity candidate pool facilitates LLM-driven reranking and precise final selection. Evaluations on the MCP-tools dataset across eight LLMs  demonstrate HGMF's superiority, outperforming over baselines like MCP-zero. Our code and data is available at \href{xxx}{xxx}.

\end{abstract}

\maketitle

\section{Introduction}
\label{sec:intro}

While Large Language Models (LLMs) demonstrate exceptional performance in language tasks, their real-world utility remains constrained without integration with external knowledge resources. 
Concurrently, the deployment of these models faces significant security and legal hurdles. 
On the security front, LLMs are susceptible to sophisticated {adversarial attacks} and latent manipulations that compromise their reliability \cite{xing2025towards, li2025optimizing, xing2025latent}. 
Meanwhile, the necessity for {intellectual property safeguards} has spurred research into robust copyright protection and tracing mechanisms to mitigate unauthorized use and data leakage \cite{xu2025evertracer, yue2025pree, xu2025copyright, xu2025rap, zhang2025meraser}.

To extend their functional boundaries, integrating external tools such as APIs and databases---a paradigm known as {tool invocation}---is essential for executing complex, multi-step operations \cite{schick2023toolformer}. 
However, a primary bottleneck in scalable tool invocation is the precise selection of candidate tools from expansive libraries. 
Presenting thousands of tool descriptions to an LLM is often infeasible due to limited context windows, semantic noise from irrelevant options, and prohibitive computational latency. 
Current solutions typically rely on preliminary filtering via keyword or vector search; however, these methods frequently overlook query nuances and the inherent {server-tool hierarchy} within libraries \cite{fei2025mcp}. 
Treating the library as a flat list often leads to suboptimal filtering, where relevant tools are discarded while noise is retained, ultimately degrading selection accuracy.

To address these limitations, we propose the {Hierarchical Gaussian Mixture Framework (HGMF)}, a probabilistic pruning framework designed for scalable and structured tool selection. 
HGMF maps the query, servers, and tools into a unified semantic space and implements a hierarchical filtering process using Gaussian Mixture Models (GMMs). 
By first identifying relevant server clusters based on query likelihood and subsequently pruning tools within those clusters, HGMF yields a refined candidate set for final reranking by an LLM.

\begin{figure*}[t]
    \centering
    \includegraphics[width=0.9\linewidth]{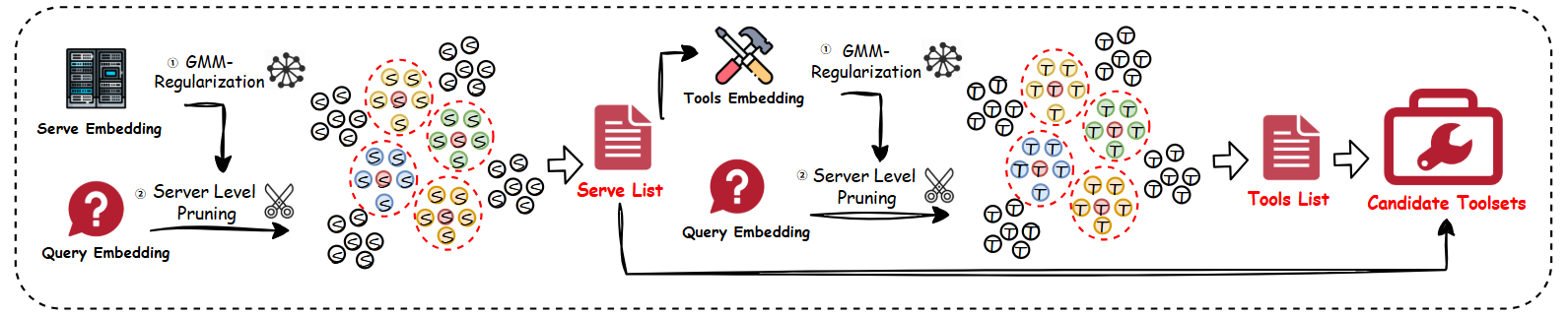}
    \caption{Overall Pipeline. The process involves embedding the user query, servers, and tools into a unified semantic space. HGMF then performs a two-level pruning strategy using clustering at the server level to obtain a relevant server list, followed by clustering at the tool level within the selected servers to yield a refined list of candidate toolsets.}
    \label{fig:overall_pipeline}
\end{figure*}

Our main contributions are:
(1) We propose \textbf{HGMF}, a novel hierarchical framework that leverages the server-tool architecture to decompose the complex search space for tool invocation.
(2) We adapt \textbf{GMMs for probabilistic soft clustering}, which more effectively captures the semantic correspondence between queries and tool clusters compared to deterministic filters.
(3) We introduce a \textbf{GMM regularization} technique, incorporating constraints such as inter-class separation and intra-class compactness, to enhance robustness and accuracy in sparse-data scenarios.
(4) We demonstrate through rigorous evaluation that HGMF achieves \textbf{state-of-the-art performance} on the MCP-tools dataset, outperforming baselines by up to 40\% in accuracy across eight LLMs while significantly reducing inference latency.

\subsection{Related Works}

\subsubsection{Tool Augmentation in Large Language Models}
Integrating external tools into LLMs has emerged as a core paradigm for transcending inherent knowledge boundaries and empowering LLMs to tackle complex real-world tasks~\cite{chen2024advancing}. From early retrieval-augmented generation (RAG)~\cite{lewis2020retrieval} to the pioneering self-supervised tool invocation in Toolformer~\cite{schick2023toolformer}, recent studies have focused on deepening LLMs' tool utilization capabilities, such as enhancing awareness of tool preconditions~\cite{yang2024can} or improving decision reliability through meta-verification and reflection learning~\cite{chen2024advancing}. Comprehensive surveys have also appeared, systematically reviewing methods and challenges in tool learning for LLM-based agents~\cite{xu2025llm}.
\subsubsection{Challenges and Existing Methods in Large-Scale Tool Selection}
Consequently, as tool libraries scale to thousands of APIs, efficient selection from vast repositories has become a central bottleneck. Standardized interfaces like the Model Context Protocol (MCP)~\cite{anthropic2024mcp} have facilitated unified interactions and ecosystem growth, yet this proliferation exacerbates the challenge by overwhelming traditional in-context learning approaches, rendering them infeasible in large-scale settings. To address this, the paradigm has shifted toward two-stage retrieval-ranking'' strategies, such as random sampling in early MCP protocols~\cite{fei2025mcp}, which yield low hit rates and prove impractical. Vector-based semantic search serves as a baseline for efficiency, but its flat'' nature introduces noise from lexical overlaps in semantically similar tools, leading to errors~\cite{fei2025mcp}. Advanced techniques, including Top-P sampling~\cite{holtzman2019curious} and dynamic retrieval~\cite{lumer2025scalemcp}, offer refinements, yet they largely neglect the inherent hierarchical structures within tool libraries.
However, these applications primarily focus on flat'' clustering of unstructured text.

\section{Methodology} \label{sec:methodology}
We propose the Hierarchical Gaussian Mixture Framework (HGMF), a three-stage process shown in Figure~\ref{fig:overall_pipeline}, to efficiently prune large toolsets for LLMs while mitigating context and noise issues. HGMF first creates semantic embeddings, then applies regularized, hierarchical GMM-based pruning, and finally uses an LLM for reranking.

\subsection{Semantic Embedding and Preprocessing} \label{ssec:embedding}

The initial stage aims to represent all textual information---the user query, server descriptions, and tool descriptions---within a unified high-dimensional semantic space. Let the set of all servers be $\mathcal{S} = \{s_1, s_2, \dots, s_M\}$ and the set of all tools be $\mathcal{T} = \{t_1, t_2, \dots, t_N\}$, where each tool $t_j$ is associated with a specific server $s_i$. Given a user query $q$, we first employ a pre-trained sentence transformer model, specifically \texttt{all-MiniLM-L6-v2}, to encode all textual descriptions into $d$-dimensional embedding vectors (where $d=384$). This process yields a query vector $\mathbf{v}_q$, a set of server vectors $\{\mathbf{v}_{s_i}\}_{i=1}^M$, and a set of tool vectors $\{\mathbf{v}_{t_j}\}_{j=1}^N$. To ensure that the subsequent similarity and probability calculations are not biased by vector magnitudes, we apply L2 normalization to all generated embeddings. For any vector $\mathbf{v}$, its normalized counterpart $\hat{\mathbf{v}}$ is computed as: $\hat{\mathbf{v}} = \frac{\mathbf{v}}{\|\mathbf{v}\|_2} \label{eq:l2_norm}$. This normalization projects all vectors onto the surface of a unit hypersphere, making cosine similarity equivalent to the dot product and stabilizing the clustering process.

\subsection{Hierarchical GMM-based Candidate Pruning} \label{ssec:clustering} 

We employ a two-level strategy using Gaussian Mixture Models (GMMs) enhanced with inter-class separation and intra-class compactness regularization. These constraints are integrated into the GMM's iterative fitting process to ensure robust and well-defined probabilistic clustering, especially on sparse data.

\paragraph{Inter-class and Intra-class Regularization}
The regularization incorporates four key parameters: inter-cluster regularization ($\lambda_{\text{inter}}$), intra-cluster regularization ($\beta_{\text{intra}}$), balance coefficient ($w_{\text{balance}}$), and covariance regularization ($\text{reg}_{\text{covar}}$). The inter-class term enforces separation by penalizing proximity between cluster centers:

\[
\mathcal{L}_{\text{inter}} = \lambda_{\text{inter}} \sum_{i=1}^{K} \sum_{j \neq i} \|\mu_i - \mu_j\|^2
\]
where \(\mu_i\) and \(\mu_j\) represent the mean vectors of the \(i\)-th and \(j\)-th cluster centers, respectively. In the case of small sample sizes, cluster centers tend to aggregate, leading to blurred cluster boundaries. Inter-class separation ensures sufficient distinction between different classes.
The intra-class regularization term is defined as:
\[
\mathcal{L}_{\text{intra}} = \beta_{\text{intra}} \sum_{i=1}^{K} \text{tr}(\Sigma_i)
\]
where \(\Sigma_i\) denotes the covariance matrix of the \(i\)-th cluster, and \(\text{tr}(\cdot)\) represents the trace of a matrix. This term maintains reasonable cluster shapes, preventing excessive elongation or deformation.
The balance term is defined as:
\[
\mathcal{L}_{\text{balance}} = w_{\text{balance}} \cdot (\mathcal{L}_{\text{inter}} + \mathcal{L}_{\text{intra}})
\]
This term balances the conflicting objectives of inter-class separation and intra-class compactness, preventing any single regularization term from dominating.

The regularized covariance matrix is defined as:
\[
\Sigma_i^{\text{reg}} = \Sigma_i + \text{reg}_{\text{covar}} \cdot I
\]
where \(I\) is the identity matrix. This regularization prevents the covariance matrix from becoming singular.
Our method incorporates regularization constraints during the M steps of Gaussian model training, followed by iterative updates until convergence. We also set convergence thresholds and maximum iteration limits to control training accuracy and duration.

\begin{table}[t]
    \centering
\caption{Average accuracy (\%) of all methods across different LLMs. The best result in each row is in \textbf{bold}. Abbreviations: Tokeniz. = Tokenization-Based, CW. = Cluster-Weighted, Density = Density-Based, Hermes. = Openhermes.}
    \label{tab:transposed_comparison}
    %\resizebox{\linewidth}{!}{%
    \footnotesize
    \begin{tabular}{l *{6}{c}}
        \toprule
        \textbf{LLM} & {\makecell{MCP-zero}} & {\makecell{HGMF}} & {\makecell{Tokeniz.}} & {\makecell{K-means}} & {\makecell{CW.}} & {\makecell{Density}} \\
        \midrule
        Mistral (7b) & 80.95 & \textbf{91.32} & 58.65 & 82.75 & 73.21 & 78.35 \\
        Gemma (7b) & \textbf{79.52} & 79.19 & 38.43 & 62.05 & 56.04 & 56.77 \\
        Llama3 (8b) & 82.74 & \textbf{84.83} & 67.50 & 83.52 & 74.73 & 83.15 \\
        Qwen (14b) & 86.26 & 87.21 & 79.88 & \textbf{87.51} & 84.25 & 78.64 \\
        Solar (10.7b) & 80.36 & \textbf{87.99} & 69.61 & 75.85 & 78.28 & 75.56 \\
        Hermes. (7b) & 75.79 & \textbf{82.57} & 47.48 & 74.15 & 71.87 & 60.87 \\
        Phi3 (3.8b) & 59.48 & \textbf{62.58} & 40.92 & 54.80 & 55.74 & 55.86 \\
        Phi4 (14b) & 79.70 & 82.99 & 80.18 & \textbf{84.95} & 82.93 & 79.60 \\
        \midrule
        \textbf{Average} & 78.10 & \textbf{82.34} & 60.32 & 75.70 & 72.13 & 71.10 \\
        \bottomrule
    \end{tabular}
    %}
\end{table}

\paragraph{Server-Level Pruning.} 
First, we model the distribution of all server embeddings $\{\hat{\mathbf{v}}_{s_i}\}_{i=1}^M$ using a GMM. A GMM assumes that the data is generated from a mixture of a finite number of Gaussian distributions with unknown parameters. We fit a GMM with $K_s$ components to the server vectors. A common heuristic, which we adopt, is to set $K_s = \lceil\sqrt{M}\rceil$ to balance model complexity and expressiveness. After fitting the model, we obtain $K_s$ Gaussian components, each characterized by a mean (cluster centroid) $\boldsymbol{\mu}_{k}$ and a covariance matrix $\boldsymbol{\Sigma}_{k}$. We then evaluate the relevance of each server cluster to the user query by calculating the likelihood of the query vector $\hat{\mathbf{v}}_q$ under each Gaussian component. The likelihood score for the $k$-th cluster is given by its probability density function: \begin{equation} \mathcal{L}(k | \hat{\mathbf{v}}_q) = \mathcal{N}(\hat{\mathbf{v}}_q | \boldsymbol{\mu}_{k}, \boldsymbol{\Sigma}_{k}) \label{eq:likelihood} \end{equation} We rank all server clusters based on these likelihood scores in descending order and select the top-$N_s$ clusters. All servers belonging to these selected clusters form our pruned server set, $\mathcal{S}' \subset \mathcal{S}$. 

\paragraph{Tool-Level Pruning.} 
Next, for each server $s_i \in \mathcal{S}'$, we consider its associated set of tool embeddings:

\[
\mathcal{T}_i = \left\{ \hat{\mathbf{v}}_{t_j} \,\middle|\, t_j \text{ is associated with } s_i \right\}
\]

We independently fit a new GMM with $K_t = \lceil\sqrt{|\mathcal{T}_i|}\rceil$ components to each tool set $\mathcal{T}_i$. Similar to the server-level pruning, we compute the likelihood of the query vector $\hat{\mathbf{v}}_q$ for each tool cluster within each selected server. For each server $s_i$, we rank its tool clusters and select the top-$N_t$ clusters. The tools belonging to these top-ranked clusters form the final pruned candidate tool set, $\mathcal{T}' \subset \mathcal{T}$. This hierarchical process ensures that only tools associated with relevant servers are considered, leading to a highly refined and contextually coherent candidate pool.

\begin{figure}[t]
    \centering
    \includegraphics[width=\linewidth]{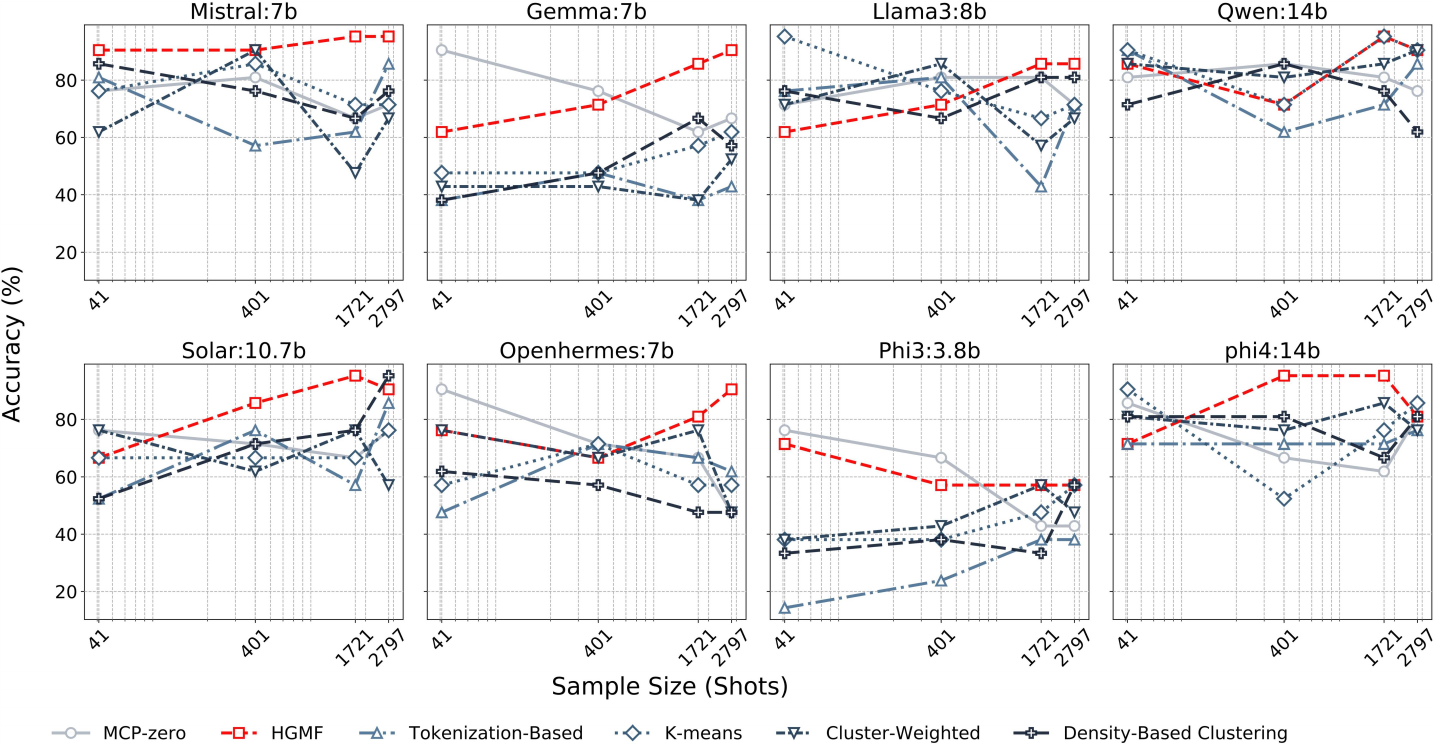}
    \caption{Performance comparison of HGMF against five baselines across eight LLMs. Each subplot shows accuracy as a function of sample size (log scale). HGMF achieves state-of-the-art results in most scenarios, showing a clear advantage at larger sample sizes.}
    \label{fig:model_comparison}
\end{figure}

\subsection{LLM-based Reranking and Final Selection} \label{ssec:reranking} 

After pruning, we have a small, structured candidate set $\{(s_i, t_j) | s_i \in \mathcal{S}', t_j \in \mathcal{T}'\}$. This set is now small enough to be efficiently processed by an LLM. 

\paragraph{LLM-driven Description Generation.} We construct a structured prompt that presents the pruned candidate set to the LLM and instructs it to act as a helpful assistant. The LLM's task is to analyze the user query $q$ and generate a natural language description of the ideal server and the ideal tool required to fulfill the request. This step leverages the LLM's superior reasoning and language understanding capabilities to synthesize a "perfect" target description based on the provided high-quality candidates.

\paragraph{Final Matching and Scoring.} Let the LLM-generated descriptions for the server and tool be $d_s^*$ and $d_t^*$, respectively. We encode these descriptions using the same sentence transformer model to obtain their embeddings, $\mathbf{v}_{s}^*$ and $\mathbf{v}_{t}^*$. We then perform the final matching by computing the cosine similarity between the LLM-generated embeddings and the embeddings of all candidates in the pruned set. The server score for a candidate server $s_i \in \mathcal{S}'$ is: $\text{score}(s_i) = \cos(\mathbf{v}_{s}^*, \hat{\mathbf{v}}_{s_i})$. Similarly, the tool similarity for a candidate tool $t_j \in \mathcal{T}'$ is: $\text{sim}(t_j) = \cos(\mathbf{v}_{t}^*, \hat{\mathbf{v}}_{t_j}) $. To determine the final ranking, we compute a combined score for each server-tool pair $(s_i, t_j)$ that accounts for both server-level and tool-level matching quality. The final score is calculated as: 
\begin{equation}
    \begin{split}
        \text{FinalScore}(s_i, t_j) ={}& (\text{score}(s_i) \times \text{sim}(t_j)) \\
                                     & \times \max(\text{score}(s_i), \text{sim}(t_j))
    \end{split}
    \label{eq:final_score}
\end{equation}

This formula gives higher weight to pairs where both the server and tool are a strong match, and it particularly rewards candidates where at least one component has an exceptionally high similarity score. The server-tool pair with the highest final score is selected as the output of our HGMF framework.

\section{Experiments}

\subsection{Experiment Configuration}

\begin{figure}[h]
    \centering
    \includegraphics[width=0.8\linewidth]{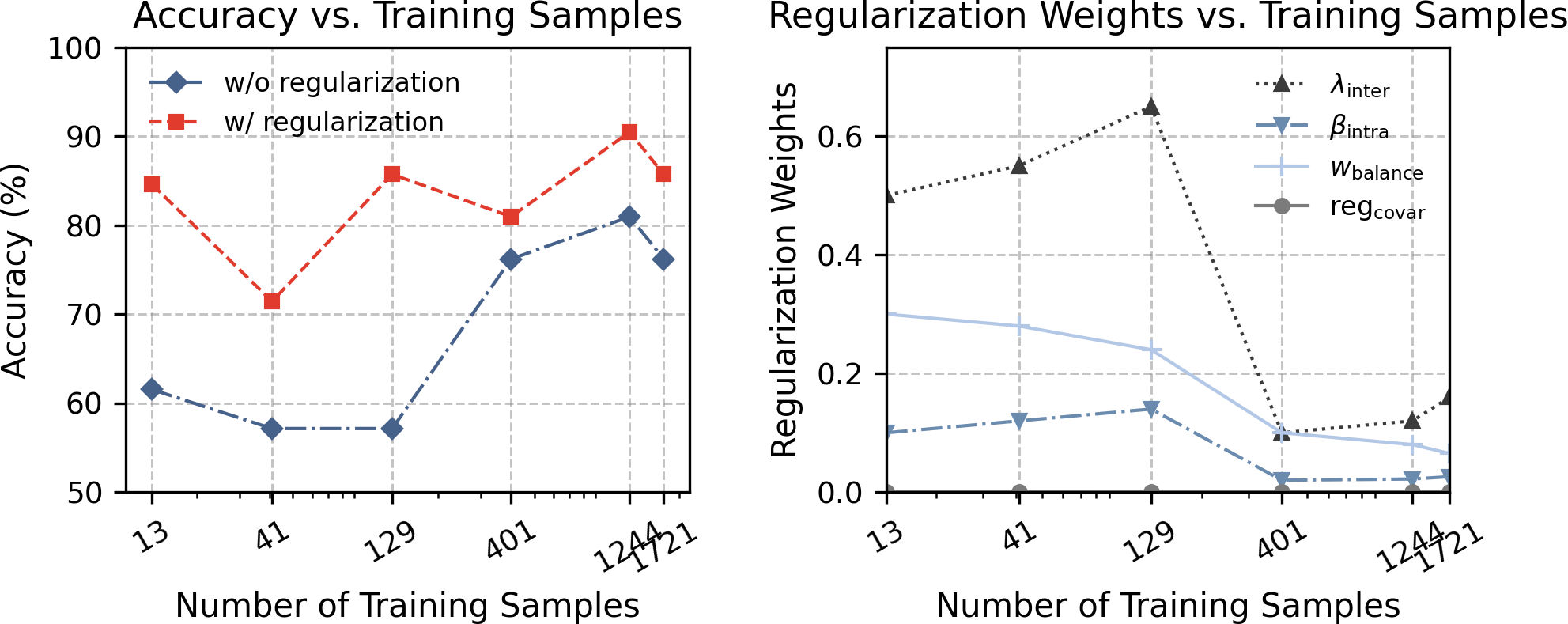}
    \caption{Accuracy comparison of HGMF with and without regularization. The regularized model achieves significant gains (14-28\%) in low-shot scenarios by mitigating cluster instability, leading to more stable and accurate performance across all sample sizes.}
    \label{fig:perform_reg}
\end{figure}

We evaluate HGMF on the MCP-tools dataset\cite{fei2025mcp}, with all text content pre-embedded using the \textbackslash{}text{all-MiniLM-L6-v2} model. The evaluation spans 10 exponential sample sizes ranging from 1 to 2,797 tools. Performance is measured by exact match accuracy (\%), where a prediction is considered correct only when both the server and tool perfectly match the ground truth. Baseline models include: MCP-zero\cite{fei2025mcp}, word-based sampling, K-means, weighted clustering, and density-based clustering.

\begin{figure}[t]
    \centering
    \includegraphics[width=0.8\linewidth]{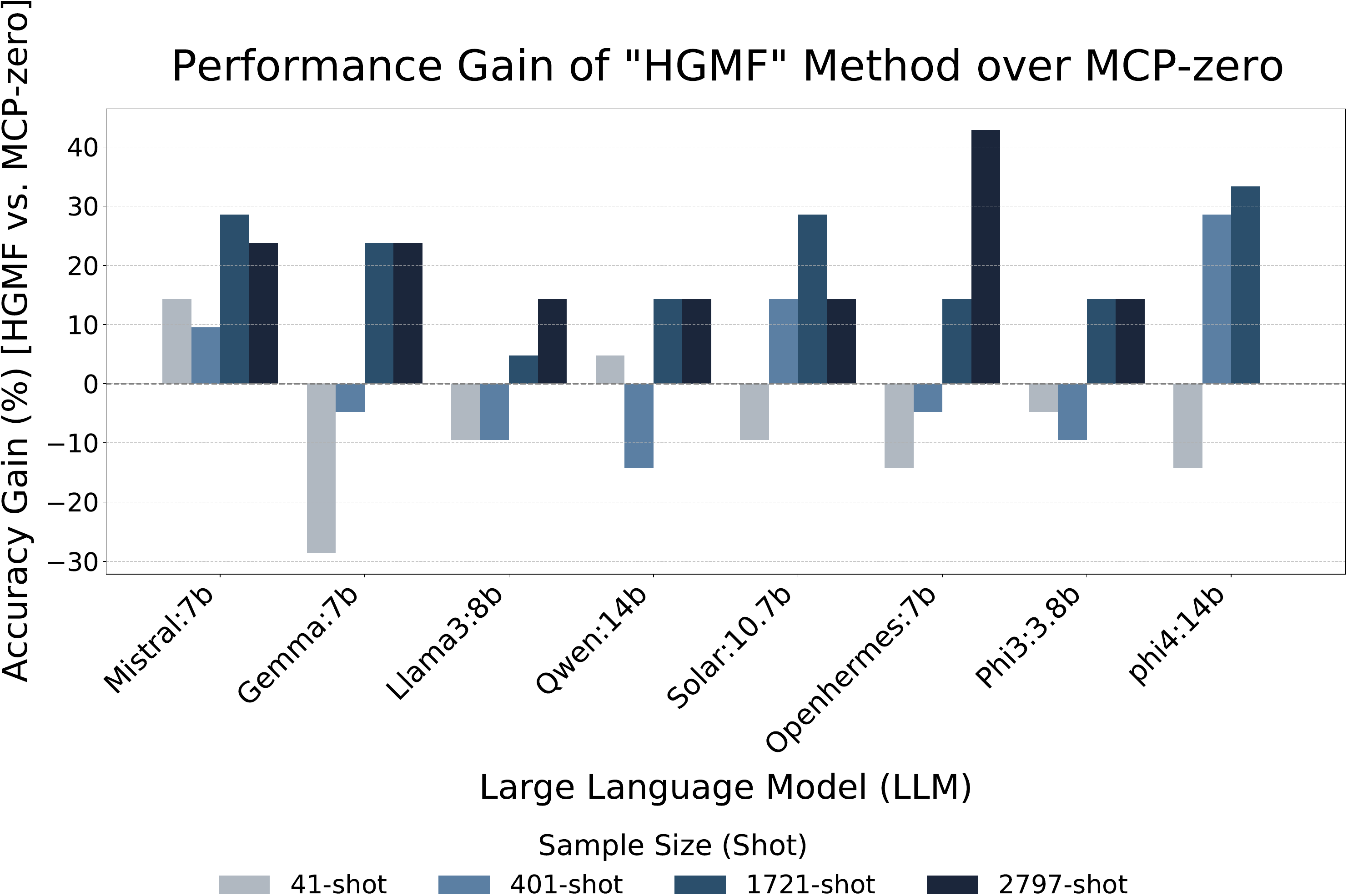}
    \caption{Accuracy gain of the HGMF method over the MCP-zero baseline. The results show that HGMF's performance advantage steadily increases with the sample size, demonstrating its superior scalability.}
    \label{fig:perform_gain}
\end{figure}

\subsection{Result Analysis}

\paragraph{Overall Performance}
Figure~\ref{fig:model_comparison} present a comprehensive comparison of our proposed method, HGMF, against five baseline methods across eight different LLMs and four sample sizes. The results demonstrate that HGMF consistently achieves superior accuracy, particularly in high-shot settings.
Figure \ref{fig:perform_gain} shows the accuracy improvement of our proposed HGMF method over the MCP-zero baseline. Positive values indicate performance gains, while negative values indicate underperformance. 
In summary, HGMF attains the highest average accuracy (82.34\%) across all models, outperforming the strong baseline MCP-zero (78.10\%) by over 4 percentage points. It achieves the best results on 6 out of 8 models. This confirms the effectiveness of our hierarchical clustering and pruning strategy in selecting high-quality tool candidates.

\paragraph{High-Shot Superiority} 
As shown in Figure\ref{fig:perform_gain}, at larger sample sizes ,HGMF consistently outperforms the baseline. It delivers significant gains—over 40 percentage points on \texttt{Openhermes:7b} and 30+ points on \texttt{phi4:14b}. highlighting its strength in distilling relevant tools.

\paragraph{Effect of Regularization}
An ablation study (Figure~\ref{fig:perform_reg}) confirms that our proposed regularization technique effectively mitigates performance instability on sparse data. While the baseline model without regularization struggles at low sample sizes (57-61\% accuracy), the fully regularized HGMF yields an average accuracy gain of 21.65 percentage points in these low-shot regimes by preventing unstable clustering.

\section{Conclusion}
To tackle context limitations and noise in Large Language Models (LLMs) selecting tools from large, hierarchical libraries, this paper introduces the Hierarchical Gaussian Mixture Framework (HGMF). It exploits the library's "server-tool" structure, using two-stage Gaussian Mixture Model (GMM) clustering for efficient probabilistic pruning to generate a compact, highly relevant candidate set for the LLM.
Experiments validate HGMF's effectiveness and scalability, achieving higher accuracy than baselines like random sampling across various LLMs. Its advantage increases with tool library size, proving suitability for large-scale, real-world toolsets. By smart filtering, HGMF enhances selection accuracy and reduces LLM inference load.
The core contribution is an efficient paradigm for large-scale tool invocation, integrating hierarchical information to break down complex selection into manageable sub-tasks. Though limited on very small toolsets, it suggests future directions like adaptive clustering parameters and extensions to graph-structured libraries.

\bibliographystyle{ACM-Reference-Format}
\bibliography{main}

\end{document}